# Fundamentals of Semantic Numeration Systems.
# Can the Context be Calculated?

Alexander Yu. Chunikhin

Palladin Institute of Biochemistry
National Academy of Sciences of Ukraine
alexchunikhin61@gmail.com

**Abstract**. This work is the first to propose the concept of a semantic numeration system (SNS) as a certain class of context-based numeration methods. The development of the SNS concept required the introduction of fundamentally new concepts such as a cardinal abstract entity, a cardinal semantic operator, a cardinal abstract object, a numeration space. The main attention is paid to the key elements of semantic numeration systems - cardinal semantic operators. A classification of semantic numeration systems is given.

## 1. Introduction

Number is a basic concept of science, an abstract entity used to describe the quantitative characteristics of objects, as well as their ordering and comparison. As we know, numbers used in practice are of two types - abstract and named (concrete).

The numeration system is a symbolic method of representing numbers using signs. Modern number systems are usually divided into three classes: positional (Place-value), non-positional and mixed. Any "place-value" numeration system deals with either abstract numbers or with uniform named numbers. Despite the significant variety of works in the field of positional numeration systems [8, 10-13], we can say that the overwhelming majority of them bear the "stamp of the game in the bases". In principle, the works on abstract numeration systems [1, 14] does not go beyond the traditionally linear representation of numbers.

Place-value representation is, in essence, the representation of an abstract number by a system of characters (digits) - the contents of places, which are named in a certain way (by numbers, symbols or words). The same applies to the representation of simple and compound named numbers.

The semantics of the traditional place-value representation can be expressed as follows: $n$ units of some abstract entity $i$ are given the meaning ($\sim$>) of a unit of another abstract entity $j$:

$$n \cdot 1_i = n_i \sim> 1_j.$$

And further, recursively:

$$m \cdot 1_j = m_j \sim> 1_k,$$
$$-------.$$

Such a representation, in which a certain amount of *one* entity is associated with a unit of *another* entity, can be called a linear or (1 - 1)-presentation.

Assume that there are such abstract entities *i*, that their *n* units $n_i$ are given the meaning of both the unit of the abstract entity *j* ($1_j$) and the unit of the abstract entity *k* ($1_k$) simultaneously: $n_i \sim> (1_j, 1_k)$. Such a representation, in which a certain amount of one entity corresponds, in the general case - different, units of several other entities, can be called a multiple, distributing or (1 - v)-representation.

Consider another situation: to form a unit of an abstract entity *k*, exactly *n* units of an abstract entity *i* and *m* units of an abstract entity *j* are required: $(n_i, m_j) \sim> 1_k$. This is a "synthetic" (2 - 1) or, in the general case, (w - 1)-representation.

Depending on how complex the system of interacting entities to be described is, operations of this kind, in various combinations, can be repeated forming a kind of semantic structure of cardinal transformations.

Can we consider this kind of structure of mutual "transformations" as a numeration system? Can the numeration systems be constructed for dissimilar quantities (numbers)? What kind of numbers can be represented in such systems? What kinds of semantic numeration systems are possible in principle?

## 2. Preliminaries

Understanding *the entity* as something distinguished in being and having meaning, we will introduce several definitions that concretize the application of this concept in the area covered.

**Definition 1**. *An abstract entity* (Æ) is an entity of arbitrary nature, provided with an identifier name that allows it to be distinguished from other entities.

For example, a number, a car, a galaxy.

The concepts of an abstract entity and an abstract object are actively discussed in modern science [7, 9]. Considering the ambiguity and variety of interpretations of these concepts, definitions adapted to the semantics of numeration systems will be given below.

Name as entity identifier can be either elementary (i) - one-element (letter, digit, word, symbol), and complex (composite, multi-element), corresponding to abstract coordinates ($<i\,|$) of the entity $Æ_{<i\,|}$ in some semantic variety. For example, "the zero bit of a number in binary notation."

Manifold (Multeity) is the next concept that is important for further presentation. From the many different definitions of this concept, we synthesize the following.

**Definition 2**. *Multeity* - the manifestation of something uniform in essence in various kinds and forms, as well as the quality or condition of being multiple or consisting of many parts.

Since in what follows we will deal with the transformation of meanings, we will define the corresponding specific type of multeity - semantic.

**Definition 3**. *Semantic multeity* is an abstract space with no more than a countable set of abstract entities, semantically united by the unity of the goal description (context).

A semantic multeity will be called *open* if the number of abstract entities in it is countable (at least potentially), *closed* if it contains a finite number of abstract entities and *bounded* if the number of abstract entities in it is finite and unchanged.
Differences between semantic multeities and manifolds in mathematics are:
- semantic *heterogeneity* of entities that make up semantic diversity;
- *goal-setting*: the semantic multeity initially includes only those abstract entities that are used (potentially can be used) to solve a specific problem;
and at the same time,
- *openness*: the possibility of replenishment (generation) of new abstract entities as a result of transformations. In the end, semantic multeity can form a semantic universe that includes any conceivable abstract entity.

**Definition 4**. *Cardinal Semantic Multeity* (CSM) is a semantic multeity, each element of which is equipped with a cardinal characteristic - the multiplicity of a given abstract entity represented in multeity.

From a set-theoretic point of view, a cardinal semantic multeity is a multiset, the carrier of which is contextually conditioned. The elements of cardinal semantic multeity will be called *cardinal abstract entities*.

**Definition 5**. *Cardinal Abstract Entity* (CÆ) is an abstract entity with a cardinal characteristic.
$$C\!\!\not{E}_i = (i; N_i),$$

where $i$ is the name of the cardinal abstract entity,
$N_i = \text{Card}(C\!\!\not{E}_i) = \#(1_i, 1_i, ..., 1_i)$, $N_i \in \mathbf{N}$.
We will assume that the named unit $1_i$ is a quantum of meaning for the abstract entity $\not{E}_i$.

## 3. The concept of the Cardinal Semantic Operator

It seems that there is no strict uniform definition of a *semantic operator* yet. Each application area - logic, linguistics, and programming - interprets this concept in its own way. However, it is possible to single out a certain semantic invariant that allows you to define the action of the semantic operator (SO) as a change / transformation of the meaning of a certain entity ($\not{E}_i$) or a set of entities ($\not{E}_i, ..., \not{E}_k$) into another meaning ($\not{E}_j$): $\not{E}_j = SO(\not{E}_i)$ or
$\not{E}_j = SO(\not{E}_i, ..., \not{E}_k)$.

Let us introduce the following basic concept - the concept of *a cardinal semantic operator*. In essence, the action of the cardinal semantic operator is to give a certain number of units $n_i$ of the cardinal abstract entity $CÆ_i$ the meaning of unit $1_j$ of the cardinal abstract entity $CÆ_j$, $(i \neq j)$: $n_i \sim> 1_j$. In principle, other options are also possible, for example, when the $n_i$ of an abstract entity $CÆ_i$ is assigned not one, but simultaneously several different semantic units of respectively different CÆs: $n_i \sim> (1_j,..., 1_k)$. Or, for the generation of the unit of meaning $1_j$ of the abstract entity $CÆ_j$, the corresponding $n$-s of several other CÆs are simultaneously needed: $(n_i,..., n_k) \sim> 1_j$.

**Definition 6**. The *cardinal semantic operator* (CSO) is a multivalued mapping of the cardinal semantic multeity CSM on itself, which associates the set of CÆ-operands $\{CÆ_i,..., CÆ_j\} \subset$ CSM with the set of CÆ-images $\{CÆ_k,..., CÆ_l\} \subset$ CSM, performing over their cardinals are operations determined by the operator signature:

$$\text{Signt(CSO)} = (K, \text{Form}, |n>_w, |r>_v),$$

where (.) is the family of the operator,
  $K$ is the kind of the operator,
  *Form* is the type of the operator,
  $|n>_w = (n_i,..., n_j)^T$ is the radix-vector, the number of components of which corresponds to the number of CÆ-operands,
  $|r>_v = (r_k,..., r_l)^T$ is a conversion vector, the number of components of which corresponds to the number of CÆ-images.

Let us consider the components of the cardinal semantic operator signature in more detail. Depending on whether the cardinal value of CÆ-operands changes under the action of the cardinal semantic operator, the latter can be representatives of one of three families - *transforming* operators (.), which change the value of the cardinals of both CÆ-operands and CÆ-images; *preserving* operators [.], changing the value of cardinals of CÆ-images and not affecting the value of cardinals of CÆ-operands; and *complex* operators (.], acting on some CÆ-operands as transforming, and on others - as preserving. This approach determines the possibility of the existence of three classes of semantic numeration systems - transforming, preserving and complex.

The kind of the cardinal semantic operator indicates the content of the transformations (definition of carry and remainder), which are performed both with the cardinals of CÆ-operands and with the cardinals of CÆ-images.

For example, the following kinds of CSOs are possible:
  - radix-multiplicity (↑#): carry $p_i = \lfloor N_i/n_i \rfloor$, remainder rem $N_i = N_i - p_i n_i$;
  - radix excess value (↑Δ): $p_i = N_i - n_i$; rem $N_i = 0$;
  - radix excess fact (↑•): $p_i = 1 \Leftrightarrow N_i > n_i$; rem $N_i = f(N_i, n_i)$;
  - arbitrary function (**f**): $p_i = f(N_i, n_i)$, rem $N_i = g(N_i, p_i, n_i)$.

In this paper, we consider the transforming cardinal semantic operators of the radix-multiplicity kind.

We will call the number of CÆ-operands of a cardinal semantic operator its *input valence* (W), $W = \#(CÆ_i,\ldots, CÆ_j) = \dim(|n>)$, and the number of CÆ-images - its *output valence* (V), $V = \#(CÆ_k,\ldots, CÆ_l) = \dim(|q>)$.

Strictly speaking, the output valence of the transforming CSO is determined by the sum of the actual output valence and the valence of "feedbacks", returning remainders to CÆ-operands. Then the full valence (–arity) of the transforming CSO will be (W, V + W). However, doubling the output valence in practice can lead to confusion, so we will usually neglect the valence of the return of reminders and write the (full) valence of the transforming CSO as (W, V).

*Radix-vector* $|n>_w = (n_i,\ldots, n_j)^T$ consists of the particular radices (bases) $n_i,\ldots, n_j$, relative to which the particular $i,\ldots, j$-carries $p_i,\ldots, p_j$ are formed. They further participate in the formation of the *common carry p* (if it's necessary).

*The conversion vector* $|r>_v = (r_{\cdot k},\ldots, r_{\cdot l})^T$ consists of the components that determine the "scale factors" of the transformation of the common carry into the components of the *transformant* $q_k,\ldots, q_l$, which change the values of the cardinals of the CÆ-images. This means that, for example, the carry *p* formed according to a given rule will be associated not with $1_j$ of the cardinal abstract entity $CÆ_j$, but with $r_{\cdot j}$ of such units. We will call $r_{\cdot j}$ the *rate of conversion* (j-conversion) of the carry. The introduction of conversion rates allows you to create numeration systems with rational bases.

The transformant $|q>_v = (q_k,\ldots, q_l)^T$ is a direct result of the action of the CSO on the CÆ-operands. However, we will consider the transformation complete only after the "recalculation" of the cardinals of all CÆ-images of the operator in accordance with the obtained values of the corresponding components of the transformant $|q>_v$.

Specific values of valency, components of the radix vector $|n>_w$ and conversion vector $|r>_v$ are determined by the *specification* of the cardinal semantic operator.

The case when the sets of CÆ-operands and CÆ-images of an operator are singleton corresponds to the generally accepted positional numeration systems.

Let us define the *main forms* of cardinal semantic operators of the radix-multiplicity kind.

1. *L-operator* (Line-operator): (↑#, L, $n_i$, $r_{ij}$) – a cardinal semantic operator of valency (W, V) = (1, 1), which assigns (gives meaning) $r_{ij}$ units of the transformant $q_j$ added to the cardinal $N_j$ of the abstract entity $CÆ_j$ to each $n_i$ of the cardinal abstract entity $CÆ_i$.

A schematic representation of the L-operator is shown in Fig.1.

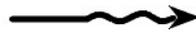

Figure 1

The functional diagram of the L-operator is shown in Fig. 2.

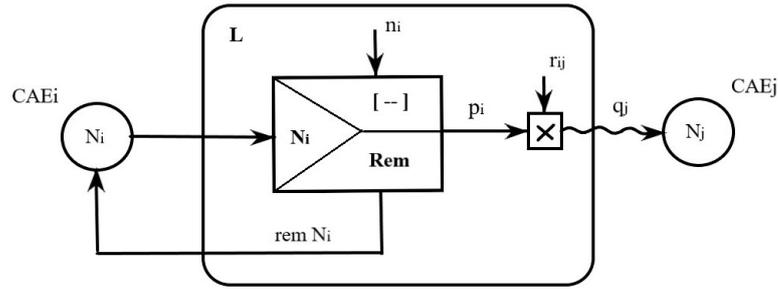

Figure 2

When an L-operator acts on a CÆ$_i$-operand, the following operations are performed:
(i)   [--]: $p_i = \lfloor N_i/n_i \rfloor$ – calculation of radix-multiplicity, that is, i-carry value;
(ii)  Rem: $N_i` = \text{rem } N_i = N_i - p_i n_i = N_i \bmod n_i$ – finding the remainder in CÆ$_i$;
(iii) $q_j = p_i \cdot r_{ij}$ – calculation of the j-transformant value;
(iv)  $N_j` = N_j + q_j$ – finding the change of the CÆ$_j$-image cardinal.

The L-operator of signature (↑#, L, $n_i$, $1_j$) is basic for many traditional positional numeration systems. In particular, for the decimal numeration system: (↑#, L, 10, 1).

2. *D-operator* (Distribution operator): (↑#, D, $n_i$, ($r_{ij}$, …, $r_{ik}$)) – a cardinal semantic operator of valency (1, v), which assigns to each $n_i$ of cardinal abstract entity CÆi $v$ transformants: $r_{ij}$ units of j-transformants $q_j$ for a cardinal abstract entity CÆj,…, $r_{ik}$ units of k-transformants $q_k$ for a cardinal abstract entity CÆ$_k$.
A schematic representation of the D-operators $D_2$ and $D_3$ is shown in Fig.3.

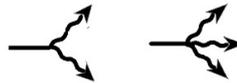

Figure 3

The functional diagram of the D-operator of valence (1, 2) is shown in Fig. 4.

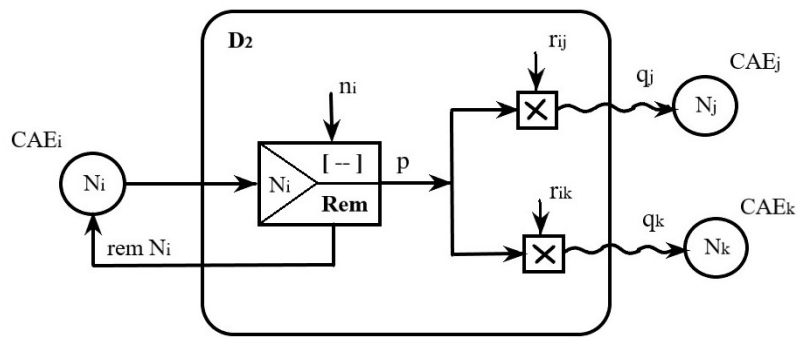

Figure 4

When a D-operator acts on a CÆ$_i$-operand, the following operations are performed:
(i) $p_i = \lfloor N_i/n_i \rfloor$ – determining radix-multiplicity (i-carry value);
(ii) $N_i` = \text{rem } N_i = N_i - p_i\, n_i = N_i \bmod n_i$ – finding the remainder in CÆ$_i$;
(iii) $q_j = p_i \cdot r_{ij}$, $q_k = p_i \cdot r_{ik}$ – calculation of partial transformants;
(iv) $N_j` = N_j + q_j$, $N_k` = N_k + q_k$ – finding the change of the CÆ-images (CÆ$_j$, CÆ$_k$) cardinals.

Note that the L-operator is a variant of the degenerate D-operator in which all $r_{ij}$, except one, are equal to zero.

3. *F-operator* (Fusion operator): (↑#, F, ($n_i$, …, $n_j$), $r_{\cdot k}$) - a cardinal semantic operator of valency (W, V) = (w, 1), which assigns to each w-tuple ($n_i$,… ,$n_j$) of CÆ-operands $r_{\cdot k}$ units of the transformant $q_k$ for the cardinal abstract entity CÆ$_k$.

A schematic representation of the F-operators $_2$F and $_w$F is shown in Fig.5.

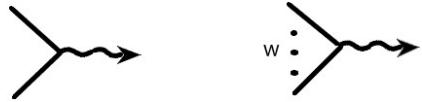

Figure 5

The functional diagram of the F-operator of valence (2, 1) $_2$F is shown in Fig. 6.

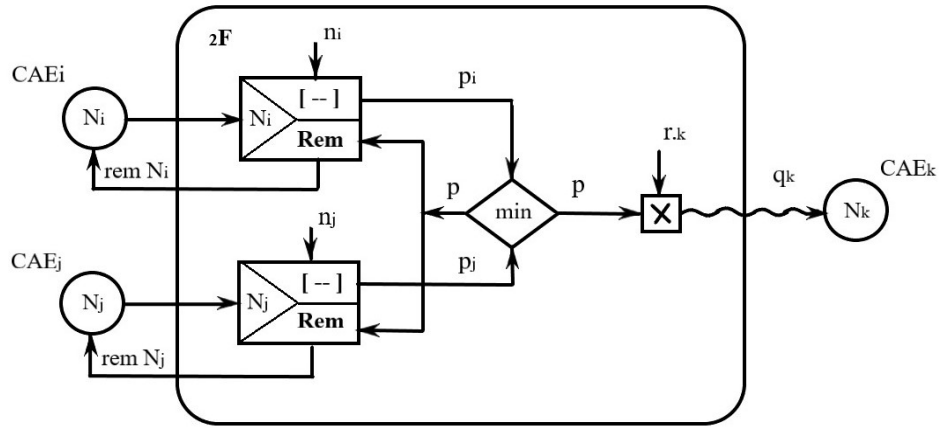

Figure 6

When a $_2$F-operator acts on CÆ$_{i,j}$-operands, the following operations are performed:
(i) $p_i = \lfloor N_i/n_i \rfloor$, $p_j = \lfloor N_j/n_j \rfloor$ – calculation of partial carries;
(ii) $p = \min\{p_i, p_j\}$ – calculation of common carry;
(iii) $N_i` = N_i - p \cdot n_i$, $N_j` = N_j - p \cdot n_j$ – calculation of the remainders in CÆ$_i$, CÆ$_j$;
(iv) $q_k = p \cdot r_{\cdot k}$ – calculation of the transformant;
(v) $N_k` = N_k + q_k$ – finding the change of the CÆ$_k$-image cardinal.

Since the partial carries $p_i, ..., p_j$ will be different, in the general case, the common carry must be determined from the condition of the existence of non-negative remainders in all CÆ-operands. This condition will be satisfied if we choose the minimal partial carry as the common carry $p$: $p = \min\{p_i, ..., p_j\}$.

4. *M-operator* (Multi-operator): $(\uparrow\#, M, (n_i,... ,n_j), (r_{\cdot k},..., r_{\cdot l}))$ - a cardinal semantic operator of valency $(W, V) = (w, v)$, which assigns to $w$-tuple $(n_i,... ,n_j)$ for CÆ-operands $v$-tuple $(r_{\cdot k},..., r_{\cdot l})$ of transformants: $r_{\cdot k}$ units of k-transformant $q_k$ for a cardinal abstract entity $CÆ_k$,..., $r_{\cdot l}$ units of l-transformants $q_l$ for a cardinal abstract entity $CÆ_l$.

A schematic representation of the M-operator $_2M_2$ is shown in Fig. 7.

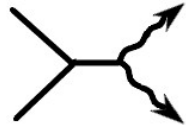

Figure 7

A functional diagram of an M-operator of valency (2, 2): $_2M_2$ is shown in Fig. 8.

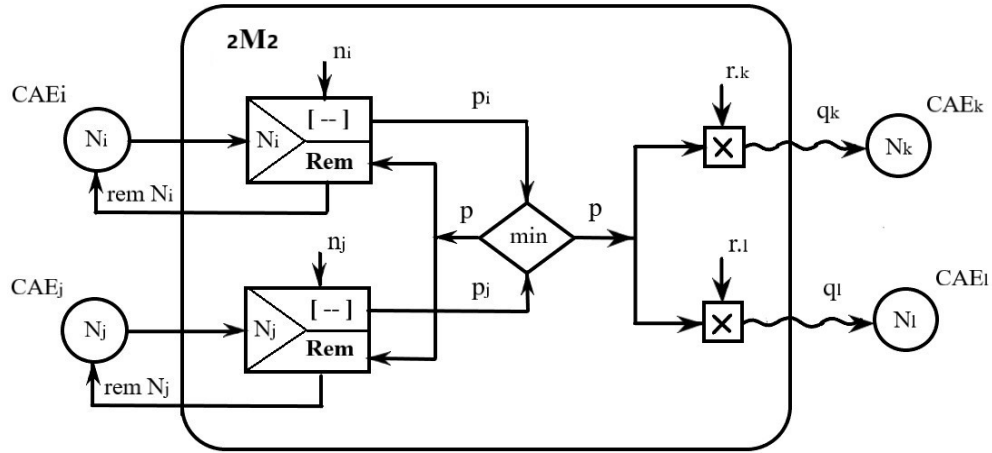

Figure 8

When the $_2M_2$-operator acts on $CÆ_{i,j}$-operands, the following operations are performed:
(i) $p_i = \lfloor N_i/n_i \rfloor$, $p_j = \lfloor N_j/n_j \rfloor$ – calculation of partial carries;
(ii) $p = \min\{p_i, p_j\}$ – calculation of common carry;
(iii) $N_i` = N_i - p \cdot n_i$, $N_j` = N_j - p \cdot n_j$ – calculation of the remainders in $CÆ_i$, $CÆ_j$;
(iv) $q_k = p \cdot r_{\cdot k}$, $q_l = p \cdot r_{\cdot l}$ – calculation of the partial transformants;
(v) $N_k` = N_k + q_k$, $N_l` = N_l + q_l$ – finding the changes of the $CÆ_{k,l}$-image cardinals.

An example of $_2M_2$-operator execution is given in Appendix 1.

It is easy to see that any of the cardinal semantic operators considered above is a special case of the M-operator: $L \sim {_1}M_1$, $D_v \sim {_1}M_v$, ${_w}F \sim {_w}M_1$. However, for the construction of specific numeration systems and the analysis of cardinal semantic transforms in them, it is often more convenient to use such reduced forms of the M-operator.

For completeness, we define two further (auxiliary) forms of operator: assignment operator $A_V$ and reset operator ${_w}Z$ (zero), as forms of M-operator. The *assignment operator* $A_V(\#_i, \ldots, \#_j)$ sets the given (usually initial) cardinal values $\#_i$ to a certain number $V$ of cardinal abstract entities in accordance with the context. The *reset operator* ${_w}Z(0_i, \ldots, 0_j)$ assigns the zero cardinal values to the cardinal abstract entities to which it applies.

**Proposition 1.** Any mono-operator cardinal semantic transformation is uniqueness.

*The proof* is based on the uniqueness of all particular mathematical transformations when performing any cardinal semantic operator.

## 4. Numeration Space. Cardinal Abstract Object

To represent complex multistage semantic transformations, mono-operator transformations, as usual, are not enough. Let us introduce the concept of a *numeration space*, the elements of which are the numeration methods. By the *method of numeration* we mean a contextually conditioned method of transforming semantic units from a cardinal semantic multeity using cardinal semantic operators. Let us formalize the last statement with the concept of a *cardinal abstract object*.

**Definition 7**. *Cardinal Abstract Object* (CAO) is a collection of cardinal abstract entities connected in a certain topology by cardinal semantic operators.

The signature of $CAO_I$:

$$Signt(CAO) = (\mathbf{I}; \mathbf{CSM}; \mathbf{CSO}; \mathbf{STop}),$$

where  **I** is the set of names denoting (naming) methods of numeration,
**CSM** is the cardinal semantic multeity,
**CSO** is the set of cardinal semantic operators,
**STop** are the possible topologies of the semantic connectivity of cardinal abstract entities by cardinal semantic operators.

**Definition 8**. *Numeration Space* (NS) is an abstract space, the elements of which are cardinal abstract objects.

A concrete $CAO_I$ implements a specific method of numeration $I$ in a numeration space.
The cardinal abstract entities included in the CAO that have an output, but do not have an input will be called *initial*, with input and output - *intermediate*, only with input - *final*, without input and output - *detached*. The main accepted assumption for the numeration methods (i.e. CAO) considered in this work is that any initial or intermediate cardinal abstract entity has a single output (associated with only one "perceiving" semantic operator) for an arbitrary (finite) number of inputs (transformants of other semantically consistent operators).

The concretizations of the CAO name, the composition of the cardinal semantic multeity, the type of operators and the topology of connectivity are determined by the *specification* of the cardinal abstract object.

## 5. Topology of Semantic Connectivity

The topology of semantic connectivity (STop) is determined by a given semantics of cardinal transformations and consists in connecting cardinal abstract entities from a cardinal semantic multeity by cardinal semantic operators of a given form.

The topology of semantic connectivity can be specified:

- descriptive/textually. For example, "serial connection of L-operators". Suitable for simple topology;
- in diagram form. For example: 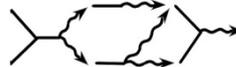

- analytically: in the form of operator formulas of various types. For example, for the above graphical representation from left to right, top to bottom: $(_2M_2|L, D_2|_2F)$;
- in tabular form.

The topology of semantic connectivity can be both "one operator"-type and "many operators"-type, both regular (for example, a 2-lattice, tree), and irregular, periodic or non-periodic, as well as cyclic. Linear topology can be specified recursively. For example, $(L)^m = L(L)^{m-1}$. An example of the complete cardinal semantic transformation of the CAO with topology of semantic connectivity $(_2M_2 \mid L, D_2 \mid _2F)$ is given in Appendix 2.

Thus, we can say that the positional numeration system in the traditional sense is a set of linearly connected cardinal abstract entities with bit semantics.

## 6. Cardinal Semantic Transformation. Multinumbers and Multicardinals

Let us agree to call a CSO "allowed" if the values of the cardinals of all its operands ensure the execution of the given operator.

**Definition 9**. *Cardinal Semantic Transformation* (CST) consists in executing, for a given $CAO_I$, all "allowed" cardinal semantic operators. A CST' *step* will mean a single execution for a given $CAO_I$ of all "allowed" cardinal semantic operators. The minimal sequence of CST' steps, leading to steady values of all cardinals in $CAO_I$, will be called a *complete* CST, and the number of such steps will be called the *length* of CST.

**Definition 10**. The multiset of cardinals of all CÆs from $CAO_I$ after an arbitrary step $\tau$ of a cardinal semantic transformations will be called the *multicardinal* of $CAO_I$ of the step $\tau$ and denoted $<A_I(\tau)>$ ($<A_I(\tau)> = [N_i(\tau), N_j(\tau), \ldots, N_k(\tau)]$).

The multicardinal meaningfully characterizes only the "card-fullness" of $CAO_I$ after each step of the cardinal semantic transformation, but in no way reflects the semantic aspect of the CST.

The multicardinal of $CAO_I$ before the first step of the transformation (CST) will be called the *initial* multicardinal $<A_I(0)>$, after a certain transformation step $\tau$ - the *intermediate* multicardinal $<A_I(\tau)>$, upon completion of transformations – the *final* multicardinal $<A_I(\omega)>$.

**Definition 11**. The holistic structural-cardinal representation of $CAO_I$ after the $\tau$-th step of the cardinal semantic transformation will be called the I-*multinumber* of the step $\tau$ (multinumber) and denoted by $A_I(\tau)$.

By analogy with the multicardinal, before the first step of CST we will call a multinumber the *initial* multinumber $A_I(0)$, after a certain step ($\tau$) of the transformation – the *intermediate* $A_I(\tau)$, upon completion of CST – the *final* multinumber $A_I(\omega)$.

An example of multinumbers of one-operator CAOs is given in Appendix 1, and an example of multinumbers and multicardinals for complete CST of a multi-operator CAO is given in Appendix 2.

We will assume that the multicardinal determines precisely the *meaning* of the CAO after the $\tau$-th step of the cardinal semantic transformation, and the multinumber is its *sense*. Informally, a multinumber is a structured multicardinal, and a multicardinal is a de-structured multinumber.

Thus, a certain I-method of numeration ($CAO_I$) is a contextually determined complete cardinal semantic transformation of both multinumbers in the numeration space (NS) and the corresponding multicardinals in the cardinal semantic multeity (CSM).

Any number represented in one or another traditional positional numeration system is a multinumber, despite the absence of cardinal semantic operators in the numbers. This is due to both the linearity of the traditionally used CSOs, and the insignificant radix variability (negotiated separately), which allows you to write numbers into a string without distorting the meaning of the representation.

## 7. Semantic Numeration Systems

Informally, the semantic numeration system (SNS) can be defined as a collection of homogeneous numeration methods.

**Definition 12**. By the *semantic numeration system* in the numeration space NS we mean its subspace $SNS_\varphi$ with the given properties determined by the classification features.

Here φ is the name-identifier of the semantic numeration system, due to a set of classification features.

SNS classification (provisional).

1. By influence on the value of the operands: *Transforming*, preserving, complex.
2. By the type of uncertainty: *Deterministic*, stochastic, fuzzy, mixed.
3. By the topology of cardinal semantic operators connectivity: *linear*, tree-like, lattice, cyclic, arbitrary (amorphous), special form.
4. By the variability of the parameters (*n* and *r*) of the cardinal semantic operators: *homogeneous* (the same for all operators), heterogeneous (different for different operators).
5. By the kind of transformation: *radix-multiplicity*, radix excess value; radix excess fact; arbitrary function; mixed.

The assignment of the semantic numeration system or its description is carried out by sequentially listing the main classification features. We recommend accepting the signs in italics as the default ones. They can be omitted when characterizing the numeration system. In the case when all features are default features, you only need to specify the radix value (as well as the value of the conversion rate, if it is different from one).

Thus, most of the generally accepted "numeration systems", for example, binary or decimal, will hardly need to be renamed. Within the framework of the above classification, they are particular (with n=2 and n=10, respectively) methods of numeration of the *transforming*, *deterministic*, *linear*, *homogeneous*, *radix-multiplicity* semantic numeration system.

Additional classifications can be introduced to characterize SNS features such as:
6. Kind of number systems used in SNS: natural, integer or rational numbers.
7. Controllability: free (autonomous) and controlled from the outside.
8. Dependence of parameters: independent (stationary) and dependent (for example,
$n_i = f(N_i)$).
9. Isotropy: constancy (isotropy) or variability (anisotropy) of the parameters of cardinal semantic operators along selected directions, for example, in a 3D lattice of topological connectivity of operators. In this case, the anisotropy can be both homogeneous and heterogeneous.

In conclusion, we state several significant *propositions*.

**Proposition 2. (The Existence)**. For any cardinal semantic multeity, there is at least one numeration method.

*The proof* is based on the possibility of linear ordering of *m* CÆs and their connection by (m-1) linear operators L. Or, for an arbitrary number *m* of cardinal abstract entities, one can form a one-operator way of numeration, for example, by operators $_{m-1}F_1$ or $_1D_{m-1}$, as well as $_{m-k}M_k$. All other options are intermediate.

**Proposition 3. (The Uniqueness)**. In any deterministic non-cyclic numeration system on a closed cardinal semantic multeity with a finite initial multicardinal $<A_I(0)>$, the final multicardinal $<A_I(\omega)>$ is unique.

*The proof* is based on the uniqueness of any mono-operator transformation (Prop.1) and on the accepted assumption that there is only one output for any non-final CÆs.

**Proposition 4. (The finiteness)**. In any deterministic non-cyclic numeration system on a closed semantic multeity, the final multicardinal $<A_I(\omega)>$ is attainable from the initial multicardinal $<A_I(0)>$ in a finite number of cardinal semantic transformation steps.

*Proof.* Since in a closed semantic multeity the number of CÆs is finite (for example, *m*), there are no cycles, and any non-final CÆ has only one output, the number of possible cardinal semantic operators can range from one to *m-1*.

## 8. Conclusion

The author's previous works [4, 5, 6] were an attempt to construct semantic numeration systems without cardinal semantic operators, but only on the basis of cardinal abstract entities connected in some topology. Abstract entities were given an active role both in the formation of carry and in the formation of a structure of connectivity (entanglement) with other abstract entities.

The semantic numeration systems considered in this article are *absolute*, where the semantic units of abstract entities characterize a purely quantitative aspect, and not the value of any quantity in a certain scale of properties.

It should also be noted that this work can serve as a certain theoretical supplement to the works [2, 3], in which the concept of an *operator* is given the status of conceptual in theory and practice.

Further work will be devoted to the development of the theory of *concrete* semantic numeration systems, i.e. numeration systems of entities, the cardinals of which are expressed not only in the absolute scale, but also in arbitrary measurement scales. It is also advisable to consider the problem of reversibility of cardinal semantic operators, the features of inverse transformations in semantic numeration systems.

**Appendix 1.** Examples of execution of cardinal semantic operators.

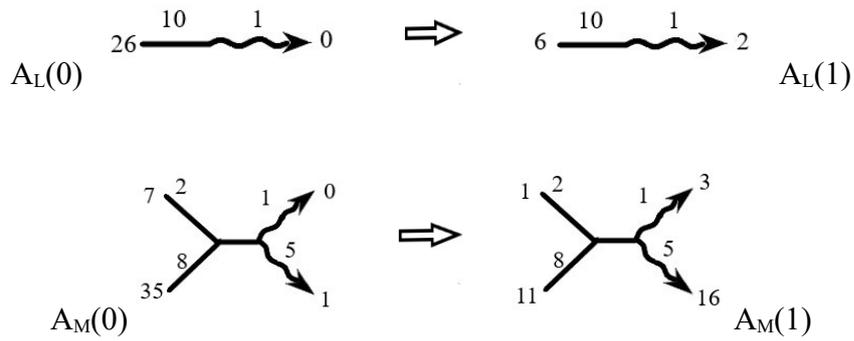

**Appendix 2.** Example of the implementation of the $CAO_I$ with topology of semantic connectivity $(_2M_2 \mid L, D_2 \mid {_2F})$. Conversion rates equal to one are not shown in the diagrams.

$A_I(0)$

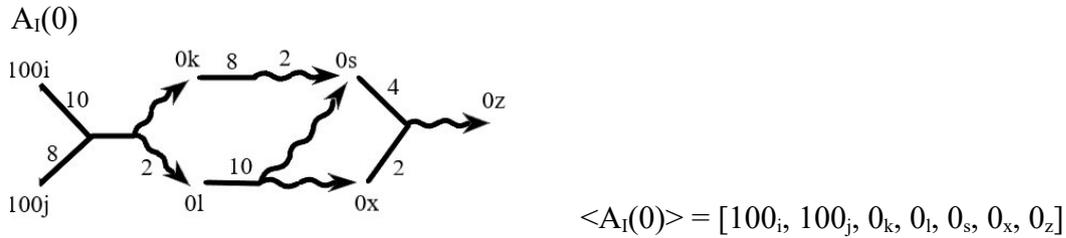

$<A_I(0)> = [100_i, 100_j, 0_k, 0_l, 0_s, 0_x, 0_z]$

$A_I(1)$

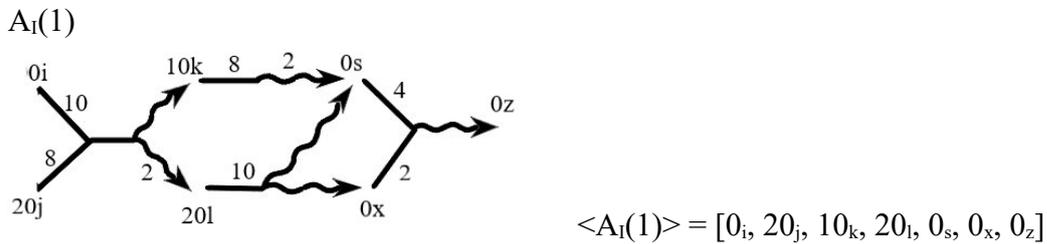

$<A_I(1)> = [0_i, 20_j, 10_k, 20_l, 0_s, 0_x, 0_z]$

$A_I(2)$

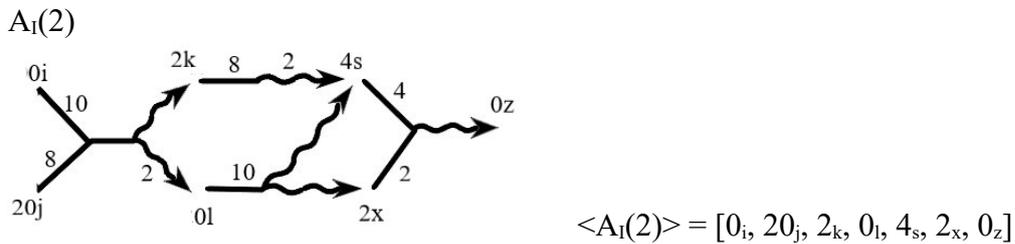

$<A_I(2)> = [0_i, 20_j, 2_k, 0_l, 4_s, 2_x, 0_z]$

$A_I(3) = A_I(\omega)$

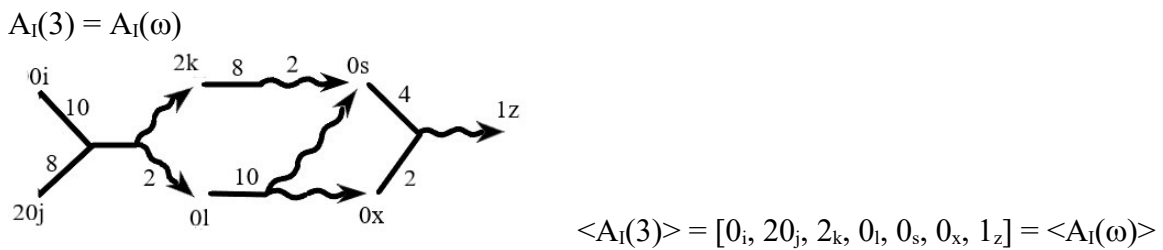

$<A_I(3)> = [0_i, 20_j, 2_k, 0_l, 0_s, 0_x, 1_z] = <A_I(\omega)>$